\documentclass[sigconf]{acmart}

\usepackage{mymacros}

\copyrightyear{2021} 
\acmYear{2021} 
\setcopyright{acmcopyright}
\acmConference[KDD '21]{Proceedings of the 27th ACM SIGKDD Conference on Knowledge Discovery and Data Mining}{August 14--18, 2021}{Virtual Event, Singapore}
\acmBooktitle{Proceedings of the 27th ACM SIGKDD Conference on Knowledge Discovery and Data Mining (KDD '21), August 14--18, 2021, Virtual Event, Singapore}
\acmPrice{15.00}
\acmDOI{10.1145/3447548.3467194}
\acmISBN{978-1-4503-8332-5/21/08}

\begin{document}
\fancyhead{}

\title{A Bayesian Approach to In-Game Win Probability in Soccer}

\author{Pieter Robberechts}
\orcid{1234-5678-9012}
\email{pieter.robberechts@kuleuven.be}
\affiliation{%
  \institution{KU Leuven, Dept. of Computer Science; Leuven.AI}
  \city{B-3000 Leuven}
  \country{Belgium}}

\author{Jan Van Haaren}
\email{jan.vanhaaren@kuleuven.be}
\affiliation{%
  \institution{KU Leuven, Dept. of Computer Science; Leuven.AI}
  \city{B-3000 Leuven}
  \country{Belgium}}

\author{Jesse Davis}
\email{jesse.davis@kuleuven.be}
\affiliation{%
  \institution{KU Leuven, Dept. of Computer Science; Leuven.AI}
  \city{B-3000 Leuven}
  \country{Belgium}
}

\renewcommand{\shortauthors}{Robberechts, et al.}

\begin{abstract}
In-game win probability models, which provide a sports team's likelihood of winning at each point in a game based on historical observations, are becoming increasingly popular. In baseball, basketball and American football, they have become important tools to enhance fan experience, to evaluate in-game decision-making, and to inform coaching decisions. While equally relevant in soccer, the adoption of these models is held back by technical challenges arising from the low-scoring nature of the sport.

In this paper, we introduce an in-game win probability model for soccer that addresses the shortcomings of existing models. First, we demonstrate that in-game win probability models for other sports struggle to provide accurate estimates for soccer, especially towards the end of a game. Second, we introduce a novel Bayesian statistical framework that estimates running win, tie and loss probabilities by leveraging a set of contextual game state features. An empirical evaluation on eight seasons of data for the top-five soccer leagues demonstrates that our framework provides well-calibrated probabilities. Furthermore, two use cases show its ability to enhance fan experience and to evaluate performance in crucial game situations.
\end{abstract}

\ccsdesc[500]{Mathematics of computing~Probabilistic inference problems}
\ccsdesc[300]{Mathematics of computing~Variational methods}
\ccsdesc[500]{Computing methodologies~Machine learning}
\ccsdesc[300]{Information systems~Data mining}

\keywords{Win probability, Soccer, Bayesian learning, Sports analytics.}

\begin{teaserfigure}
  \centering
  \includegraphics[width=.85\textwidth]{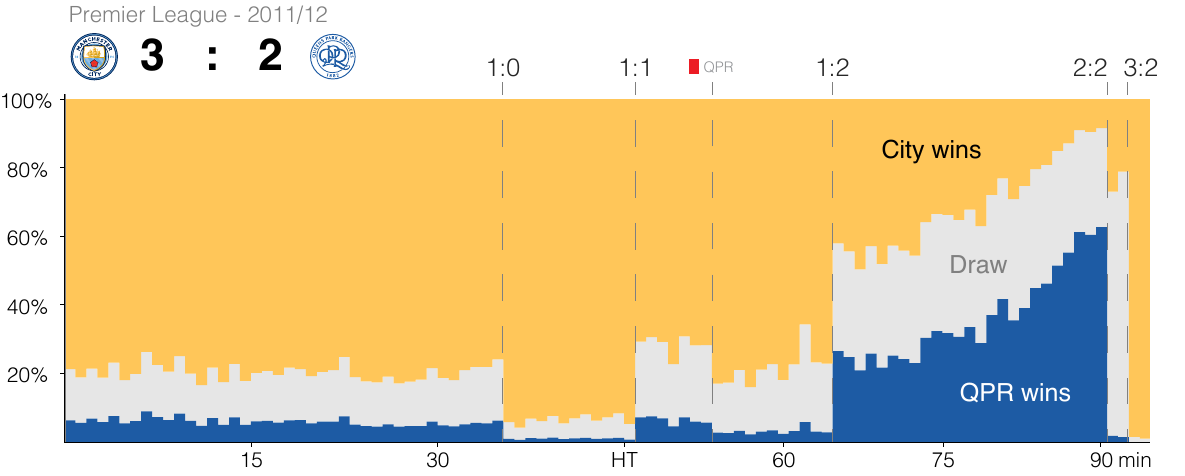}
  \caption{In-game win probabilities for the game between Manchester City and Queens Park Rangers (QPR) on the final day of the 2011/12 Premier League season.}
  \Description{In-game win probabilities for the game between Manchester City and Queens Park Rangers.}
  \label{fig:teaser}
\end{teaserfigure}

\maketitle

\section{Introduction}


Soccer outcome prediction has been the subject of research since at least the 1960s~\cite{reep1968}. While prior research has always focused on predicting the probability that a particular team will win the game prior to kickoff, this paper introduces an in-game win probability model that continuously updates the probability throughout the course of the match based upon the evolving game conditions (i.e., score, time remaining, etc.). Figure~\ref{fig:teaser} illustrates our model's output by showing the minute-by-minute win/tie/loss probabilities during the iconic game between Manchester City and Queens Park Rangers (QPR) on the final day of the 2011/12 English Premier League season,	where Manchester City needed a win to secure the title. 
The evolution of the probabilities illustrates how City started the game as the clear favorite, ended up in dire straits in the second half after QPR took the lead in the 66\textsuperscript{th} minute, and made their riveting injury time comeback.  

While this example illustrates the value of in-game win probability models as a "story stat", they also have a multitude of other valuable applications in sports analytics and betting. First, whenever any rational analysis is performed on any sport, one is almost inevitably concerned with the impact of either performance or strategy on the probability of winning. Hence, the win probability added (WPA) metric --- computed as the change in win probability between two consecutive game states --- is a popular metric to rate a player's contribution to his team's performance~\cite{pettigrew2015,lindholm2014}, measure the risk-reward balance of coaching decisions~\cite{lock2014,morris2017}, or evaluate in-game decision making~\cite{mcfarlane2019}. Second, win probability changes can help identify exciting or influential moments in the game~\cite{schneider2014}, which may be useful for broadcasters looking for game highlights~\cite{decroos2017} or to evaluate performance in crucial game situations~\cite{robberechts2019,pettigrew2015}. Finally, they are relevant to in-game betting scenarios. Here, gamblers have the option to continue to bet once a sports event has started, and adapt their bets depending on how the event is progressing. 

Nowadays, in-game win probability is widely used as a statistical tool in baseball~\cite{lindsey1961}, basketball~\cite{beuoy2015,ganguly2018} and American football~\cite{burke2010,pelechrinis2017,lock2014}.
In contrast, it is a relatively new concept in soccer, where it first emerged during the 2018 World Cup when both FiveThirtyEight~\cite{fivethirtyeight2018} and Google published such predictions. In December 2019, win probabilities provided by Opta debuted during live broadcasting~\cite{opta2019}. Unfortunately, none of these companies provide any details about how they implemented their models. 

The lack of publicly documented win probability models can likely be attributed to the unique challenges that come with modeling these probabilities for soccer. In particular, it involves challenges such as capturing the current game state, dealing with stoppage time, the frequent occurrence of ties and changes in momentum.  


To address these challenges, this paper introduces a Bayesian in-game win probability model for soccer that models the future number of goals that each team will score as a temporal stochastic process. We train and evaluate our model on eight seasons of event stream data from the top five European soccer leagues. Finally, we introduce two relevant use cases for our win probability metric: a ``story stat'' to enhance the fan experience and a tool to quantify player performance in the crucial moments of a game with applications in player recruitment.

\section{Task and Challenges}
\label{sec:task}

The task of an in-game win probability model is to predict the probability distribution over the possible match outcomes\footnote{For simplicity, we only consider the outcome at the end of regulation time in this paper. Nevertheless, the proposed model can easily be extended to handle the possibility of overtime and penalties in the knockout phase of tournaments.}  $P(Y |\ x_{t})$ given the game state $x_{t}$ at time $t$. 
This problem has been solved for baseball, basketball and American football, where generally one of the following four modelling techniques is used:

\begin{enumerate}
    \item \textbf{Cumulative historical average~\cite{pettigrew2015,albert2007}} A na\"{i}ve model may simply report the cumulative historical average for a very compact game state representation (e.g., score differential and time remaining).
    \item \textbf{Logistic regression classifier (LR)~\cite{pelechrinis2017,burke2010}.}  When more detailed game state representations are used, one can use a basic logistic regression model to generalize from previously observed states to ones that have never been seen before:
    \begin{equation}
    P(Y | x_{t}) = \frac{e^{\Vec{w}^T \vec{x}_t}}{1 + e^{\Vec{w}^T \Vec{x}_t}},
    \end{equation}
    where Y is the dependent random variable of the model representing whether the game ends in a win, tie or loss for the home team, $\Vec{x}_t$ is the vector with the game state features, while the coefficient vector $\Vec{w}$ includes the weights for each independent variable and is estimated using historic match data.
    \item \textbf{Multiple logistic regression classifiers (mLR)~\cite{burke2009,torvik2017}.}
    This model removes the remaining time from the game state vector and trains a separate logistic classifier per time frame. As such, this model can deal with the non-linear effects of the time remaining on the win probability. 
    \item \textbf{Random forest model (RF)~\cite{lock2014}.} Finally, a random forest model can deal with non-linear interactions between all game state variables.
\end{enumerate}

It seems straightforward that a similar model could be applied to soccer too. Yet, soccer has some unique distinguishing properties that impact developing a win probability model. We identify four such issues:

\textbf{1. Describing the game state.}
In-game win probability models are based on features that capture the estimated pre-game strength difference between the two teams, as well as features that describe the in-game situation. The former are well researched in pre-game prediction models~\cite{hvattum2010}. The latter are unique to in-game prediction models. For each sport, they include at least the time remaining and the score differential, but the remaining features that describe the in-game situation differ widely between sports. American football models typically use features such as the current down, distance to the goal line and number of remaining timeouts~\cite{pelechrinis2017,lock2014}, while basketball models incorporate possession and lineup encodings~\cite{beuoy2015,ganguly2018}. These contextual features mainly have their value in tight matches, when there is a small score difference between two teams. The importance of these features rapidly diminishes as soon as the score difference increases. In contrast, in soccer, large score differences are far less common and teams may take the lead when being outplayed due to a lucky goal. Hence, describing the game state is more challenging for soccer. 

\textbf{2. Dealing with stoppage time.}
In most sports, one always knows exactly how much time is left in the game, but this is not the case for soccer. Soccer games rarely last precisely 90 minutes. Each half is 45 minutes long, but the referee can supplement those allotted periods to compensate for stoppages during the game. There are general recommendations and best practices that allow fans to project broadly the amount of time added at the end of a half, but no one can be certain about the amount of added time until it is announced just prior to the end of a half.

\textbf{3. The frequent occurrence of ties.}
Predicting the outcome of soccer games based on an in-game scenario is a non-standard predictive task due to the frequent occurrence of ties. In basketball and American football, the possibility of ties can be ignored due to their rare occurrence. Therefore, win probability in these sports can be modelled with standard binary classifiers. Although these classifiers can easily be adapted to a multinomial setting, it is desirable to take the ordering nature of the win-draw-loss outcome in soccer into account to prevent losing valuable information. So far, the ordered probit or ordered logit model are the workhorse models in such cases, but it is apparent that the relation between the remaining time, current score difference and win-draw-loss outcome is non-linear. However, there has been little working exploring non-linear classifiers for ordered categorical outcomes.



\textbf{4. Changes in momentum.}
Additionally, the fact that goals are scarce 
means that when they do occur, they often change the subsequent ebb and flow of the game in terms of how space opens up and who dominates the ball -- and where they do it. The existing in-game win probability models are very unresponsive to such shifts in the tenor of a game. 

\section{In-game Win Probability Framework}
In this section, we outline our approach for constructing an in-game win probability model for soccer. First, we introduce our model, which addresses the challenges described in the previous section. Second, we discuss how to represent the game state. Finally, we describe our training and evaluation procedures.

\subsection{Modelling match outcomes}

Most existing in-game win probability models use a machine learning model that directly estimates the probability of the home team winning. Instead, we model the number of future goals that a team will score and then map that back to the win-draw-loss probability. Specifically, given the game state at time $t$, we model the probability distribution over the number of goals each team will score between time $t+1$ and the end of the match. 
This task can be formalized as: 

\begin{problem_def}
  \probleminput{A game state $(x_{t, home},\ x_{t, away})$ at time $t$.}
  \problemquestion{}
    \problemdetails{Estimate probabilities
    \begin{itemize}
        \item $P(y_{>t, home} = g\ |\ x_{t, home})$ that the home team will score $g \in \mathbb{N}$ more goals before the end of the game
        \item $P(y_{>t, away} = g\ |\ x_{t, away})$ that the away team will score $g \in \mathbb{N}$ more goals before the end of the game
    \end{itemize}
    such that we can predict the most likely final scoreline $(y_{home},\ y_{away})$ for each time frame in the game as 
    $
        (y_{<t, home}+y_{>t, home},\ y_{<t, away} + y_{>t, away})
    $.
  }
\end{problem_def}

This formulation has two important advantages. First, the goal difference contains a lot of information and the distribution over possible goal differences provides a natural measure of prediction uncertainty~\cite{ganguly2018}. By estimating the likelihood of each possible path to a win-draw-loss outcome, our model can capture the uncertainty of the win-draw-loss outcome in close games. Second, by modeling the number of future goals instead of the total score at the end of the game, our model can better cope with these changes in momentum that often happen after scoring a goal. 

We model the expected number of goals that the home  ($y_{>t, home}$) and away ($y_{>t, away}$) team will score after time $t$, as independent Poisson distributions:
\begin{equation*}
    \begin{aligned}
        &y_{>t,\mathrm{home}} &\sim \textrm{Pois}((T-t) * \theta_{t,\mathrm{home}}), \\
        &y_{>t,\mathrm{away}} &\sim \textrm{Pois}((T-t) * \theta_{t,\mathrm{away}}),
    \end{aligned}
\end{equation*}

\noindent where the $\theta$ parameters represent each team's estimated scoring intensity in the $t$\textsuperscript{th} time frame. Although several researchers~\cite{lee1997, karlis2003} have shown the existence of a (relatively low) correlation between the number of goals scored by the two opponents, this has been ignored  since it demands more sophisticated techniques.


The scoring intensities are estimated independently from the game state features $x_{t}$ of the home and away team. The importance of these game state features varies over time. At the start of the game, the prior estimated strengths of each team are most informative, while near the end the features that reflect the in-game performance become more important. Moreover, this variation is not linear, for example, because of a game's final sprint. Therefore, we model these scoring intensity parameters as a temporal stochastic process. In contrast to a multiple regression approach (i.e., a separate model for each time frame), the stochastic process view allows sharing information and performing coherent inference between time frames. As such, our model can make accurate predictions for events that occur rarely (e.g., a red card in the first minute of the game). More formally, we model the scoring intensities as:

\begin{equation*}
\begin{aligned}[c]
\theta_{t, home} &= \invlogit (\Vec{\alpha}_t * x_{t, home} + \beta  + \textrm{Ha}) \\
\theta_{t, away} &= \invlogit (\Vec{\alpha}_t * x_{t, away} + \beta) 
\end{aligned}
\end{equation*}
\begin{equation*}
\alpha_0 \sim N(0,2) ,\quad
\alpha_t \sim N(\alpha_{t-1},2) ,\quad
\beta \sim N(0,2) ,\quad
\textrm{Ha} \sim N(0,2)
\end{equation*}
where $\Vec{\alpha}_t$ are the time-varying regression coefficients, $\beta$ is the intercept and $\textrm{Ha}$ models the home advantage. We choose zero-mean normal priors with variance of two for the regression coefficients and home advantage, which corresponds to weak information regarding the true parameter values.

\subsection{Describing the game state}
\label{sec:features}
The variable duration of games in soccer hampers a straightforward implementation of this framework. If the game clock would be used directly to determine the current time frame $t$, the number of time frames would vary between games such that $t=45$ would correspond to the end of the first half in some games, and up to 8 minutes before half time in other games. 
To deal with the variable duration of games due to stoppage time, we split each game into $T=100$ time frames, each corresponding to a percentage of the game. Halftime always corresponds to $T=50$. 

Next, we describe the game state in each of these frames for each team separately using the following variables:
\begin{enumerate}
    \item \textbf{Base features}
    {
    \begin{itemize}
        \item Game Time: Percentage of game time completed.
        \item Score Differential: The current score differential.
     \end{itemize}
     }
 \item \textbf{Team strength features}
     {
     \begin{itemize}
         \item Rating Differential: The difference in Elo ratings~\cite{hvattum2010} between both teams, which represents the prior estimated difference in strength with the opponent.
    \end{itemize}
    }
 \item \textbf{Contextual features}
 {
 \begin{itemize}
    \item Team Goals: The number of goals scored so far.
    \item Reds: The difference with the opposing team in the number of red cards received. A double yellow card is counted as a red card.
    \item Yellows: The number of yellow cards received by the opponent. A second yellow card is not counted.
    \item Goal-Scoring Opportunities: The number of goal-scoring opportunities per time frame that a team created so far, averaged by the number of time frames already passed. These opportunities include successful shots, blocked shots and situations where a player was in a good position to score.
    \item Attacking Passes: A rolling average of the number of successfully completed attacking passes (a forward pass ending in the final third of the field) during the previous 10 time frames.
    \item Expected Threat (xT)~\cite{karun2019,vanroy2019}: A rolling average of a team's xT during the previous 12 time frames. This metric rewards moving the ball into ``threatening'' positions that, in turn, have a high probability of leading to good shooting positions. 
    \item Duel Strength: A rolling average of  the percentage of duels won in the previous 10 time frames.
  \end{itemize}
  }
\end{enumerate}

The challenge here is to design a good set of contextual features. The addition of each variable increases the size of the state space exponentially and makes learning a well-calibrated model significantly harder. On the other hand, they should accurately capture the likelihood of each team to score goals in the remaining game time. The seven contextual features that we propose are capable of doing this: the number of goals scored so far gives an indication of whether a team was able to score in the past (and is therefore probably capable of doing it again); a difference in red cards represents a goal-scoring advantage~\cite{cerveny2018}; a weaker team that is forced to defend can be expected to commit more fouls and incur more yellow cards; the number of goal-scoring opportunities, the percentage of successful attacking passes and xT captures a team's momentum in creating goal-scoring opportunities; and finally, the percentage of duels won captures how effective teams are at regaining possession. 

\subsection{Model training}
Since the complex Bayesian model introduced in the previous section cannot be solved with analytical approaches, we require some form of a Probabilistic Programming Language (PPL) to estimate the posterior distributions of its parameters. Typically, PPLs implement Markov Chain Monte Carlo (MCMC) algorithms that allow one to draw samples and make inferences from the posterior distribution implied by the choice of model conditional on the observed data. However, MCMC algorithms are computationally expensive and do not scale very easily. That makes MCMC algorithms unfit for our problem, since we need large amounts of data to infer the likelihood of each match outcome in an almost limitless state space.


Over the past few years, however, a new class of algorithms for inferring Bayesian models has been developed, that do not rely heavily on computationally expensive random sampling. These algorithms are referred to as Variational Inference (VI) algorithms and have been shown to be successful with the potential to scale to ``large'' datasets~\cite{blei2017}. Hence, our model was trained using PyMC3's\footnote{We used PyMC3 v3.8. See \url{https://docs.pymc.io/}.} Auto-Differentiation Variational Inference (ADVI) algorithm~\cite{kucukelbir2017}. We also take the opportunity to make use of PyMC3’s ability to compute ADVI using ``batched'' data, which further facilitates model training at scale.


\subsection{Evaluation procedures}
An in-game win probability model should provide calibrated probability estimates that reflect what is most likely to happen in reality. For example, when a team is given an 8\% probability of winning at a given state of the game, this essentially means that if the game was played from that state onward a hundred times, the team is expected to win approximately eight of them. This cannot be assessed for a single game, since each game is played only once, but we can collect all game states for which our model predicts an 8\% home win probability and then look at whether about 8\% of the corresponding games actually resulted in a home win. We validate the model's calibration using both reliability diagrams~\cite{niculescu2005} and the multi-class expected calibration error (ECE)~\cite{guo2017}. For obtaining these, we distribute the predicted win-draw-loss probabilities into $M$ bins and compute the average difference between the predicted and expected outcomes, weighted by the number of examples in each bin. The ECE is derived from these bins as:
\begin{align}
   \textrm{ECE} = \sum_{m=1}^M\frac{|{B_{m}|}}{N}\left| \bar{y}(B_m) - \bar{p}(B_m) \right|,
\end{align}
where $N$ is the number of samples, $B_m$ is the $m$-th probability bin, $|B_m|$ denotes the size of the bin, and $\bar{p}(B_m)$ and $\bar{y}(B_m)$ denote the average predicted probability of the most likely outcome and the proportion of positives in bin $B_i$ respectively. In the experiments, we use M = 5.

Besides calibration, it is also important to quantify how close the predictions are to the actual outcome.
Therefore, we use the Ranked Probability Score (RPS)~\cite{constantinou2012} at match time $t$: 
\begin{equation}
    RPS_t = \frac{1}{2} \sum_{i=1}^2(\sum_{j=1}^i p_{t,j} - \sum_{j=1}^i e_j)^2,
    \label{eq:RPS}
\end{equation}
where $\Vec{p}_{t} = [P(Y = \text{win}\ |\ x_t),\ P(Y = \text{tie}\ |\ x_t),\ P(Y = \text{loss}\ |\ x_t)]$ are the estimated probabilities at a time frame $t$ and $\Vec{e}$ encodes the final outcome of the game as a win ($\Vec{e} = [1,1,1]$), a tie ($\Vec{e}  = [0, 1, 1]$) or a loss ($\Vec{e}  = [0,0,1]$). 
This metric reflects that an away win is in a sense closer to a draw than a home win. That means that a higher probability predicted for a draw is considered better than a higher probability for home win if the actual result is an away win.
\section{Experiments}
The goal of our experimental evaluation is to: (1) explore the prediction accuracy and compare with the LR, mLR and RF modelling approaches, (2) evaluate the importance of each feature and (3) identify the challenges for real-time deployment.

\subsection{Dataset}
Our analysis relies on event stream data from the English Premier League, Spanish LaLiga, German Bundesliga, Italian Serie A and French Ligue 1. This data was scraped from \url{http://whoscored.com}. For each league, we used the 2011/12 up to 2017/18 seasons to train our models. This training set consists of 12,758 games. The 2018/19 season of each league was set aside as a test set containing 1,826 games. Due to the home advantage, the distribution between wins, ties and losses is unbalanced. In the full dataset, 46\% of the games end in a win for the home team, 25\% end in a tie and 29\% end in a win for the away team.

To assess the pre-game strength of each team, we scraped Elo ratings from \url{http://clubelo.com}. For soccer, the single rating difference between two teams is a highly significant predictor of match outcomes~\cite{hvattum2010,robberechts2018}. 

\subsection{Model evaluation}
\begin{figure}[!h]
    \centering
    
    \begin{subfigure}[t]{.45\linewidth}
    \end{subfigure}
    \hfill
    \begin{subfigure}[t]{.45\linewidth}
        \includegraphics[width=\linewidth]{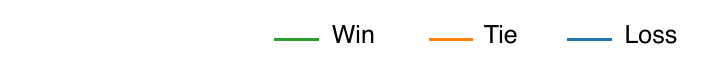}
    \end{subfigure}
    \hfill
    
    \begin{subfigure}[t]{.45\linewidth}
        \includegraphics[width=\linewidth]{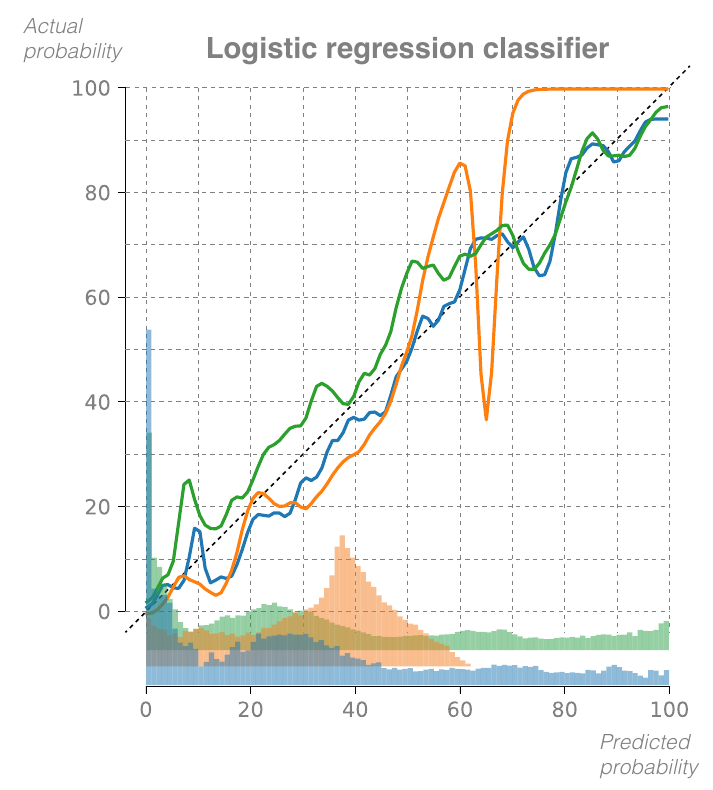}
    \end{subfigure}
    \hfill
    \begin{subfigure}[t]{.45\linewidth}
        \includegraphics[width=\linewidth]{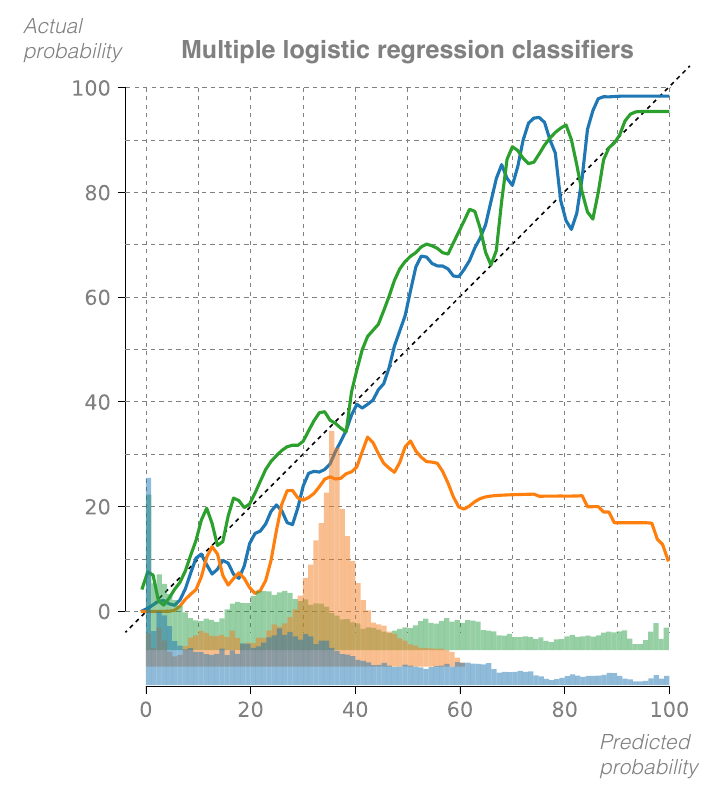}
    \end{subfigure}
    
    \begin{subfigure}[b]{.45\linewidth}
        \includegraphics[width=\linewidth]{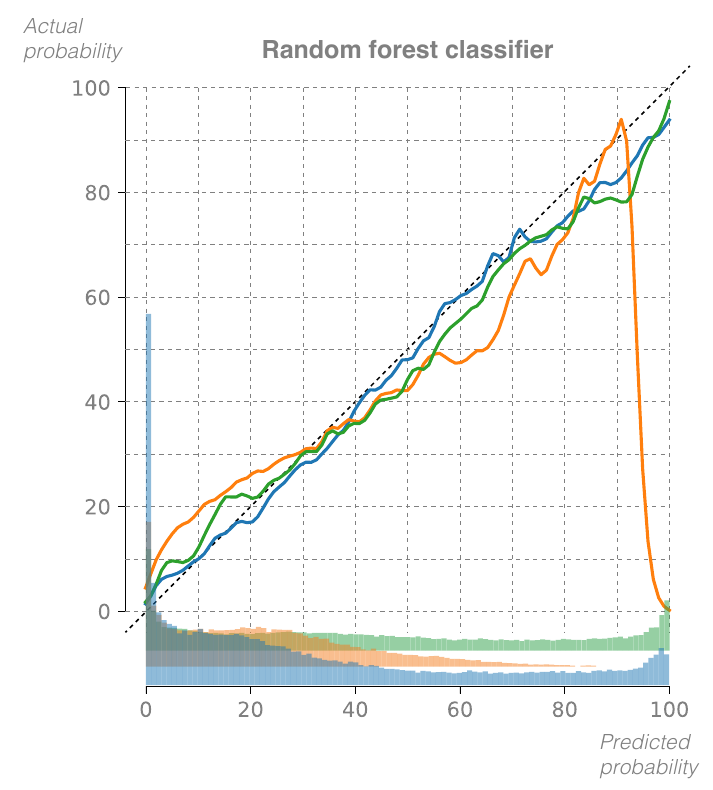}
    \end{subfigure}
    \hfill
    \begin{subfigure}[b]{.45\linewidth}
        \includegraphics[width=\linewidth]{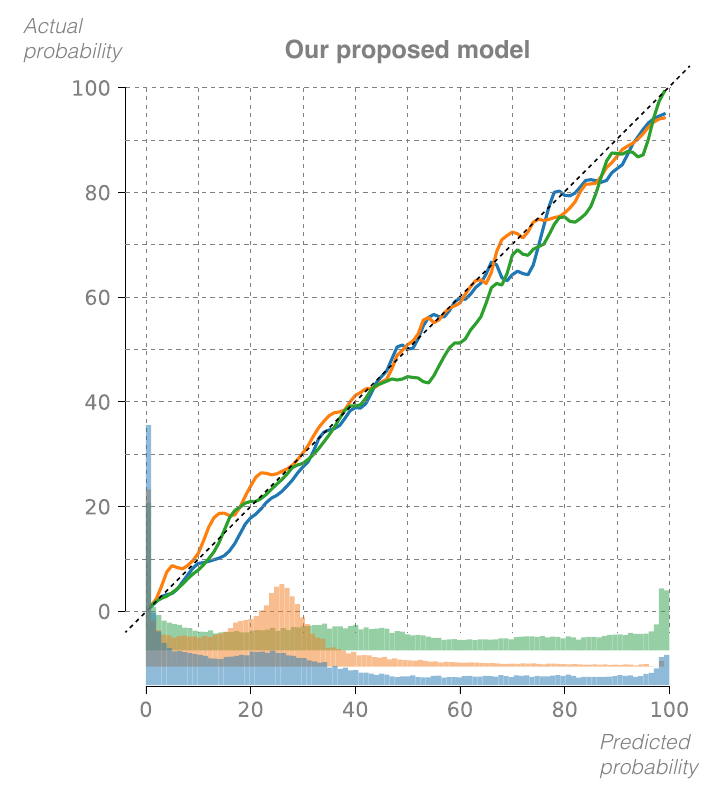}
    \end{subfigure}
    \caption{Probability calibration curves and histograms of the predicted probabilities for the four considered approaches. Only the Bayesian classifier has well calibrated win, draw and loss probabilities.}
    \label{fig:pc_existing_models}
\end{figure}

Figure~\ref{fig:pc_existing_models} compares the calibration aggregated over the entire duration of a game of our proposed Bayesian model against the approaches used in other sports. Table~\ref{table:calibration} gives the ECE for these same models, aggregated separately for each half and for the final 10\% of the game. All models were trained with the same set of features. 

\begin{table}[ht]
\caption{Expected calibration error (ECE) of logistic regression (LR), multiple logistic regression (mLR), random forest (RF) and our proposed classifier. In addition to the overall ECE, we provide the ECE for the first and second half, as well as for the final ten percent of games. Note that each time window corresponds to different data. Hence, it is only possible to make comparisons of the ECE among the models within the time window.}
\label{table:calibration}
\begin{tabular}{@{}lrrrr@{}}
\toprule
               & \multicolumn{1}{c}{H1} & \multicolumn{1}{c}{H2} & \multicolumn{1}{c}{Final 10\%} & \multicolumn{1}{c}{Overall} \\
\midrule
LR             & 0.031 & 0.069 & 0.174 & 0.023 \\
mLR            & 0.023 & 0.067 & 0.170 & 0.027 \\
RF             & 0.051 & {\bf 0.013} & 0.101 & 0.024 \\
Proposed model & {\bf 0.012} & {\bf 0.013} & {\bf 0.002} & {\bf 0.011} \\
\bottomrule
\end{tabular}
\end{table}

Only our Bayesian model has good probability calibration curves. The difference is most apparent in the final 10\% of games, when our proposed Bayesian model achieves an ECE of 0.002. Among the three other models, the RF classifier performs well, but the predictions for the probability of ties break down in late-game situations (ECE = 0.101). It never predicts a high probability of a tie. Similarly, the LR and mLR models struggle to accurately predict the probability of ties. Additionally, their win and loss probabilities are also not well calibrated. 

\begin{figure}[b]
    \centering
    \includegraphics[width=.8\linewidth]{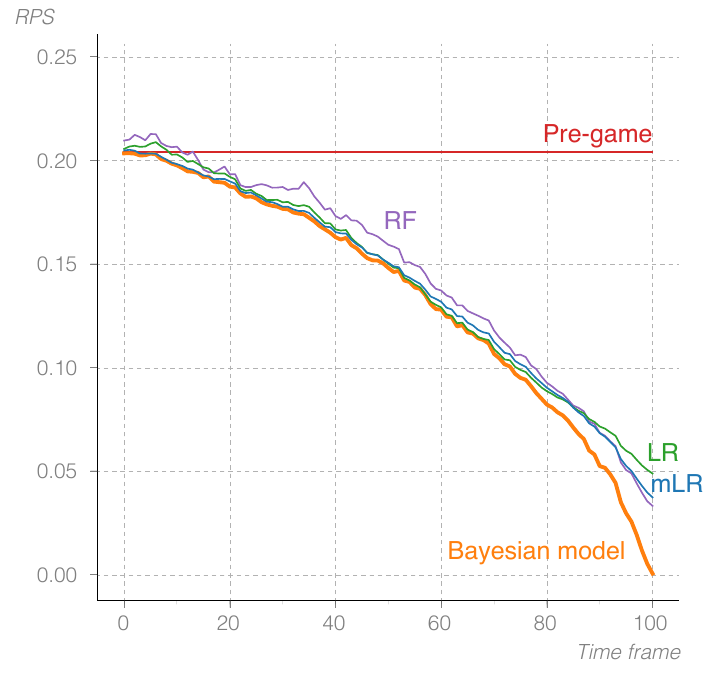}
    \caption{All models' performance improves as the game progresses, but only our Bayesian model makes consistently correct predictions at the end of each game. Early in the game, the performance of all models is similar to an Elo-based pre-game win probability model.}
    \label{fig:rps_all_models}
\end{figure}

The RPS of all in-game win probability models improves when the game progresses (Figure~\ref{fig:rps_all_models}), as they gain more information about the final outcome. Yet, only our Bayesian model is able to make consistently correct predictions at the end of each game. For the first few time frames of each game, the models' performance is similar to a pre-game logistic regression model that uses the Elo rating difference as a single feature. Furthermore, our Bayesian model clearly outperforms the LR, mLR and RF models.


\subsection{Feature importance}
\begin{figure}[!h]
    \centering
    \includegraphics[width=.8\linewidth]{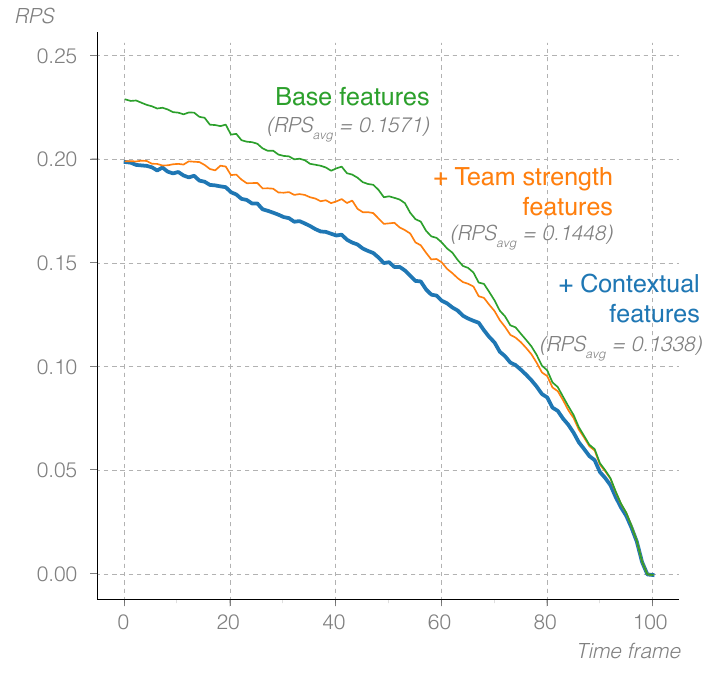}
    \caption{The RPS of the win probability model improves by adding team strength features and contextual features.}
    \label{fig:rps_features}
\end{figure}

In addition to the predictive accuracy of our win probability estimates, it is interesting to observe how these estimates are affected by the different features. To this end, Figure~\ref{fig:rps_features} shows the RPS of our model trained with (1) only the base features (game time and score differential), (2) base features and team strength features, and (3) base features, team strength features and contextual features. Each feature group further improves the RPS. The addition of pre-game strength features has the largest positive impact on prediction accuracy early in the game. However, as soon as the game starts, prediction accuracy significantly improves due to the addition of contextual features. For the last 10\% of a game, the added value of team strength and contextual features is minimal. During the final moments of the game, the current score differential is a good predictor for the final match outcome.

\begin{figure}[!h]
    \centering
    \includegraphics[width=\linewidth]{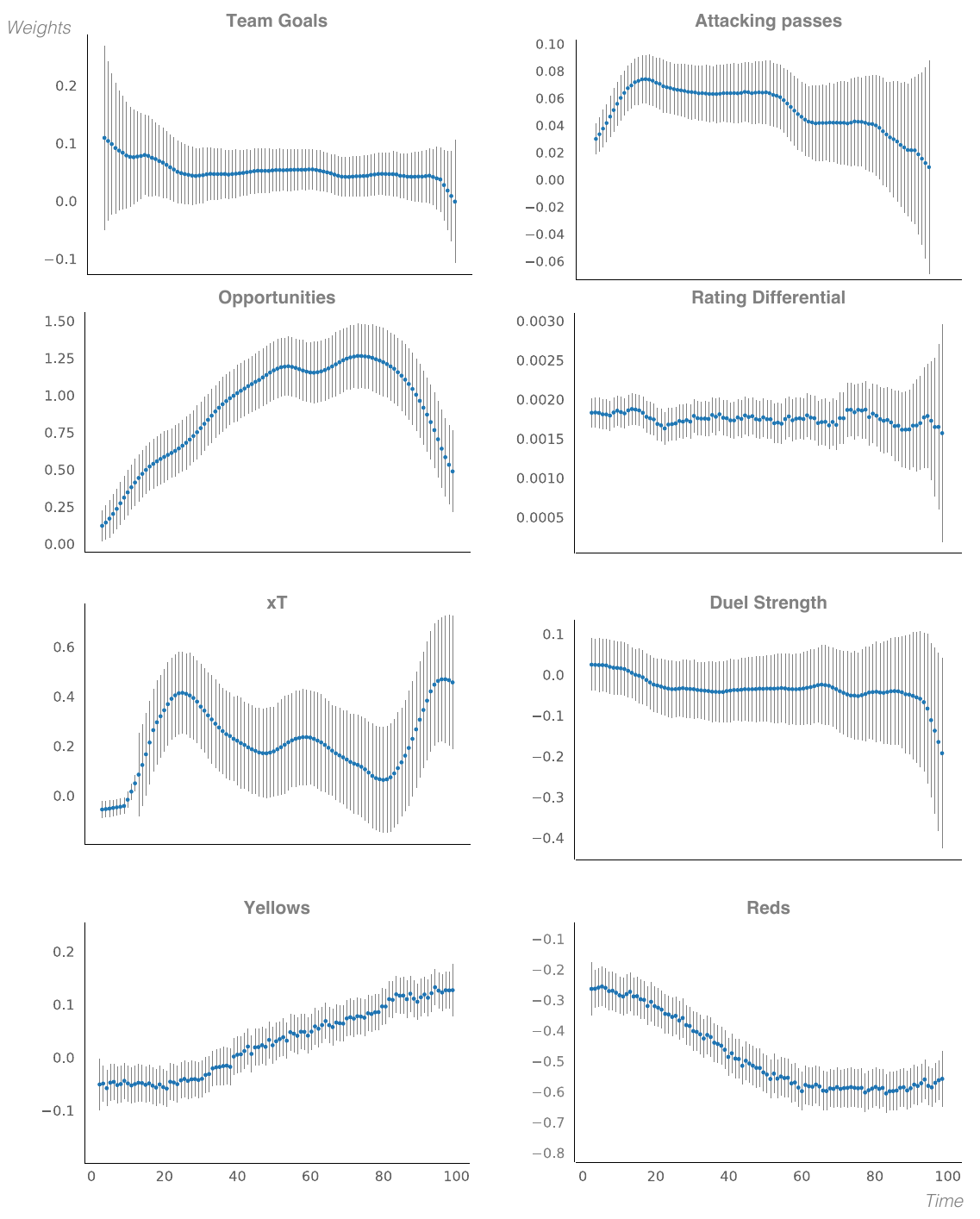}
    \caption{Estimated mean weight and variance for each feature per time frame.}
    \label{fig:feat_importance}
\end{figure}

It is also interesting to look at the weights that are given to each individual feature and how these weights change over the course of a game. To this end, we inspect the simulated traces of the weight vector $\vec{\alpha}$ for each feature. In the probabilistic framework, these traces form a marginal distribution on the feature weights for each time frame. Figure~\ref{fig:feat_importance} shows the mean and variance of these distributions. Primarily of note is that winning more duels has a negative effect on the win probability, which is not what one would intuitively expect. Yet, this is not a novel insight.\footnote{\url{https://fivethirtyeight.com/features/what-analytics-can-teach-us-about-the-beautiful-game/}} Furthermore, we notice that a higher Elo rating than the opponent, previously scored goals, yellow cards for the opponent, more successful attacking passes, the creation of goal-scoring opportunities and a higher xT value all have a positive impact on the scoring rate. On the other hand, receiving red cards decreases a team's scoring rate. Finally, the effect of yellows increases as the game progresses. Attacking passes, xT and goal-scoring opportunities seem to interact with each other around the 80th time frame. The importance of the xT feature increases, while the other two features lose importance. Since xT is more fine-grained compared to the other features, it is probably better at capturing a team's offensive intentions during the last part of a game. Furthermore, red cards and duel strength have a bigger impact on the scoring rate in the first half. For the pre-game Elo rating differential, mainly the uncertainty about the effect on the scoring rate increases during the game. 

\subsection{Challenges for real-time deployment}

There are two challenges for deploying our framework to make real-time predictions. First, the time that will be added at the end is unknown for unfinished games.  On average, a soccer match is 96 minutes long, with two minutes of added time in the first half and four minutes of added time in the second half. One can estimate this added time using a random forest model based on the current game time, the number of substitutes, the number of yellow and red cards, the number of goals scored, the time lost due to injuries, the league and whether the game is tense (goal difference less than or equal to one). Using a separate model for each half, we obtain mean absolute errors of 46.2s in the first half and 59.4s in the second half.


Second, four of our contextual features (attacking passes, xT, goal-scoring opportunities and duel strength) require detailed event stream data. Providers like Opta deliver these streams in near real-time,\footnote{\url{https://www.optasports.com/services/data-feeds/}} but at a high cost and with limited accuracy. Alternatively, a simplified model without these three features could be used for real-time predictions, which would slightly reduce the RPS from 0.134 for the offline model to 0.138 for the real-time model. 

\section{Use Cases}
In-game win-probability models have a number of interesting use cases. In this section, we first show how win probability can be used as a story stat to enhance fan engagement. Second, we discuss how win probability models can be used as a tool to quantify the performance of players in the crucial moments of a game. We illustrate this with an Added Goal Value (AGV) metric, which improves upon standard goal scoring statistics by accounting for the value each goal adds to the team's probability of winning the game.

\subsection{Fan engagement}
As Figure~\ref{fig:teaser} illustrates for Manchester City's iconic 3-2 victory over QPR, win probability is a great ``story stat'' because it provides historical context to specific in-game situations and illustrates how a game unfolded.
Undoubtedly fans implicitly considered these win probabilities as the game unfolded. However, where football fans and commentators have to rely on their intuition and limited experience, win probability stats can deliver a more objective view. Therefore, win probability could be of interest to fans and commentators to quantify the back-and-forth that occurred during a particularly exciting game and to put the (un)likeliness of certain game situations into a historical context.

Providing well-calibrated probability estimates is crucial to establish trust in the model. This applies in particular to the final moments of the game, when the shorter window makes it easier for people to grasp the likelihood of each outcome. Additionally, given that people most often look at win probability charts after a major comeback, a model that provides unreasonable predictions in these situations is unlikely to be adopted by the media.

\subsection{Quantifying clutch performance}
Another interesting application of in-game win probability is its ability to identify the most crucial moments of a game, since these can be regarded as situations where scoring or conceding a goal would have a large impact on the expected outcome of the game~\cite{robberechts2019}. This enables a new set of metrics to measure ``clutch'' performance, or performance in crucial situations, which is a recurring concept in many sports --- including soccer.

To illustrate this idea, we show how win probability can be used to identify clutch goal scorers in soccer. The number of goals scored is the most important statistic for offensive players. Yet, soccer talent scouts recognize that not all goals are created equal. Scoring the winning goal in stoppage time, for example, is regarded as being more difficult and clearly more valuable than another goal when the lead is already unbridgeable. By using the change in win probability\footnote{We remove the pre-game strength from our win probability model for this analysis. Otherwise, games of teams such as PSG that dominate their league would all start with an already high win probability, reducing a goal's impact on the win probability.} when a goal is scored, we can evaluate how much a player's goal contributions impact their team's chance of winning the game. This leads to the Added Goal Value metric, similar to Pettigrew's~\cite{pettigrew2015}:
\begin{equation*}
       \textrm{AGVp90}_i = \frac{\sum_{k=1}^{K_i} 3 * \Delta P(\text{win} | x_{t_k}) + \Delta P(\text{tie} | x_{t_k}) }{M_i} * 90,
\end{equation*}
where $K_i$ is the number of goals scored by player $i$, $M_i$ is the number of minutes played by that same player and $t_k$ is the time at which a goal $k$ is scored.


This formula calculates the total added value that occurred from each of player $i$'s goals, averaged over the number of games played. Since both a win and a draw can be an advantageous outcome in football, we compute the added value as the sum of the change in win probability multiplied by three and the change in draw probability. The result can be interpreted as the average boost in expected league points that a team receives each game from a player's goals.

\begin{figure}[h]
    \centering
    \includegraphics[width=\linewidth]{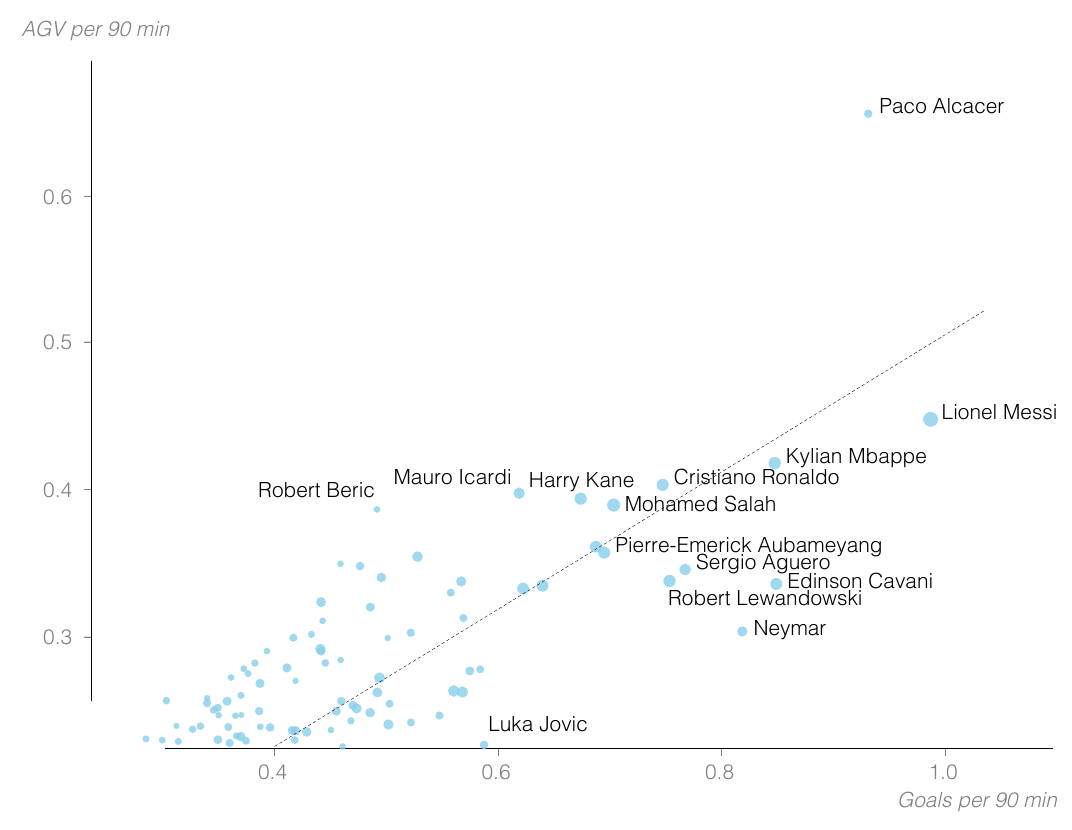}
    \caption{The relation between goals scored per 90 minutes and AGVp90 for the most productive Bundesliga, Ligue 1, Premier League, LaLiga and Serie A players in the 2017/2018 and 2018/2019 seasons.}
    \label{fig:AGVp90}
\end{figure}

Figure~\ref{fig:AGVp90} displays the relationship between AGVp90 and goals per game for the most productive Bundesliga, Ligue 1, Premier League, LaLiga and Serie A players who have played at least the equivalent of 20 games and scored at least 10 goals in the 2017/2018 and 2018/2019 seasons. The players with the highest AGVp90 are Paco Alc\'acer, Lionel Messi and Kylian Mbapp\'e. The diagonal line denotes the average AGVp90 for a player with similar offensive productivity. Players below this line such as Neymar, Robert Lewandowski and Edinson Cavani have a relatively low added value per goal; while players above the line such as Robert Beric, Harry Kane and Paco Alc\'acer add more value per goal than the average player. What strikes one most is the exceptionally high AGVp90 of Paco Alc\'acer. After his move to Dortmund during the summer of 2018, Alc\'acer scored 12 out of his 18 goals in the 80th minute or later. This tally has directly earned Dortmund twelve points.

AGVpg90 will partially capture players who are able to score when playing against a top team or in a tie game. Hence, adding the AGVp90 metric to SciSports' player recruitment platform\footnote{\url{http://platform.scisports.com/}} would enable  scouts to identify attackers who perform well in situations of increased mental pressure. Most goal-based soccer statistics count each goal equally, whereas the AGVp90 metric weighs more important goals higher than less important goals. As a result, this metric is particularly useful for scouts at soccer clubs that play many tense games while fighting for the championship or battling against relegation.


\section{Related Work}
In-game win probability models emerged for Major League Baseball in the 1960s~\cite{lindsey1961}. Given the relatively low number of distinct game-states and the large dataset of historical games, it is possible to accurately predict the probability of winning based on the cumulative historical average final result for a given game state.

A similar approach has been used in ice hockey -- which has a similar score progression compared to soccer -- albeit using a very limited game state representation~\cite{pettigrew2015,moneypuck}. These models estimate the win probability from a historical average for a given score differential and time remaining, but also take into account power plays by calculating the probability that a goal will be scored based on the time left in the penalty, and then incorporate the effect that such a goal would have on the win expectancy.

More detailed game state representations have successfully been used in win probability models for American football and basketball. These models typically use some regression approach to generalize from previously experienced states to ones that have never been seen~\cite{burke2010,pelechrinis2017,lock2014,beuoy2015,torvik2017}.

An alternative approach to in-game win probability prediction is to estimate the final scoreline by simulating the remainder of the game. For example,~\citet{rosenheck2017} developed a model that considers the strength of offense and defense, the current score differentials and the field position to forecasts the result of any possession in the NFL. This model can be used to simulate the rest of the game several times to obtain the current win probability.~\citet{vstrumbelj2012} did something similar for basketball. These approaches can be simplified by estimating the scoring rate during each phase of the game instead of looking at individual possessions. For example,~\citet{buttrey2011} estimate the rates at which NHL teams score using the offensive and defensive strength of the teams playing, the home-ice advantage, and the manpower situation. The probabilities of the different possible outcomes of any game are then given by a Poisson process. Similarly,~\citet{stern1994} assumes that the progress of scores is guided by a Brownian motion process instead, and applied this idea to basketball and baseball. These last two approaches are most similar to the model that we propose for soccer. In contrast to these methods, we consider the in-game state while estimating the scoring rates and assume that these rates change throughout the game and throughout the season. 

Our proposed in-game win probability model belongs to the class of time-varying coefficient models~\cite{hastie1993,franco2019}. These models have been used in a wide range of other applications~\cite{canova1993,dangl2012}, but are novel for the prediction of in-game win probability. 
\section{Conclusions}
This paper introduced a Bayesian in-game win probability model for soccer. The model uses ten features for each team and models the future number of goals that a team will score as a temporal stochastic process. Empirically, the predictions made by this model are well calibrated and outperform the typical modelling approaches that are used in other sports.
The model has relevant applications in sports story telling and can form a central component for analyzing player performance in the crucial moments of a game.

\begin{acks}
We thank SciSports for facilitating this research. This research received funding from the Flemish Government (AI Research Program). Pieter Robberechts is supported by the KU Leuven Research Fund (C14/17/070).
\end{acks}

\bibliographystyle{ACM-Reference-Format}
\balance
\bibliography{main}
\appendix

\section{Experimental Setup and Implementation}
We performed all experiments in this paper in Python. We evaluated
the performance of four learning approaches, for which we provide details below:
\begin{description}
\item[Logistic Regression] We used the implementation from the scikit-learn\footnote{\url{https://scikit-learn.org/stable/modules/generated/sklearn.linear_model.LogisticRegression.html}} Python package. We used an L2 regularization penalty, the stochastic average gradient (SAG) method for solving the optimization problem, adjust weights inversely proportional to class frequencies in the input data, and use a maximum of 1000 iterations for the solver to converge. Using a 5-fold cross-validation grid search for the regularization constant \texttt{C: [0.01, 0.1, 1, 10, 100]}, we found that the default value of 1 gave the best result.
\item[Multiple Logistic Regression] We used the same implementation and parameters settings as for the pooled logistic regression model above.
\item[Random Forest] We used the implementation from the scikit-learn\footnote{\label{RF}\url{https://scikit-learn.org/stable/modules/generated/sklearn.ensemble.RandomForestClassifier.html}} Python package to train a forest of 100 trees with a maximum tree depth of 100, considering a maximum of 3 features for each split, requiring a minimum of 3 samples at each leaf node and a minimum of 8 samples to split an internal node. We identified these parameter settings using a 5-fold cross-validation grid search with  
\begin{lstlisting}
    'max_depth': [80, 90, 100, 110],
    'max_features': [2, 3],
    'min_samples_leaf': [3, 4, 5],
    'min_samples_split': [8, 10, 12],
    'n_estimators': [100, 200, 300, 1000]
\end{lstlisting}
\item[Bayesian Temporal Stochastic Process] We used PyMC's (v3.8) variational inference API\footnote{\url{https://docs.pymc.io/api/inference.html?highlight=advi\#pymc3.variational.inference.ADVI}} with 200,000 iterations and mini-batches of 500 games. For evaluation, we take the average of 2,000 posterior predictive samples.
\end{description}

We trained and evaluated all models on a computing server running Ubuntu 18.04 with 32GB of RAM and an Intel(R) Xeon(R) CPU E3-1225 v3 @ 3.20GHz. The runtime for training the Bayesian model is approximately 76 minutes.

\subsection{Predicting Stoppage Time}
To enable real-time predictions, we use two separate random forest models for each half to estimate the time that will be added at the end. Both models use the same set of features:
\begin{itemize}
    \item The current game time in minutes
    \item The number of substitutes 
    \item The number of yellow cards for both teams
    \item The number of red cards for both teams
    \item The number of goals scored by both teams
    \item The time lost due to injuries and time-wasting, measured as
    \begin{itemize}
        \item The number of seconds between two successive events for all periods with no event for each pair of two successive events with a gap of at least 30 seconds
        \item The number of periods with no event for a period of at least 1, 2, 4 and 8 minutes
    \end{itemize}
    \item A league identifier
    \item Whether the goal difference is less than or equal to one (i.e., whether the game is tense)
\end{itemize}
We trained both models on the 2016/17 and 2017/18 seasons of the top-5 European leagues and evaluate on the 2018/19 seasons. We used the random forest implementation from the scikit-learn\footnotemark[\getrefnumber{RF}] Python package to train a forest of 20 trees with a maximum tree depth of 5, and requiring a minimum of 20 samples at each leaf node. The results are shown in Table~\ref{tab:est_stoppage_time} for predictions made at the end of the regular play time of the first and second period.

\begin{table}[ht]
\caption{Mean absolute error (MAE) and $R^2$ score for the estimation of stoppage time.}
      \centering
        \label{tab:est_stoppage_time}
        \begin{tabular}{@{}lrr@{}}
        \toprule
                          & \textbf{MAE} & \textbf{$R^2$} \\ \midrule
        \textbf{Period 1} & 46.2s        & 0.41                 \\
        \textbf{Period 2} & 59.4s        & 0.24                 \\ \bottomrule
        \end{tabular}
\end{table}


\section{Alternative Game State Features}
Besides the final set of game state features which we described above, we considered a large list of alternative features listed below. These features are based on expert knowledge and the decision on which features to include the model was based on multiple experiments with varying combinations of features, varying aggregation methods (sum or rolling average) and feature parameters (number of past time frames to consider). \\

\noindent\textbf{Offensive strength}
\begin{itemize}
    \item Offensive strength: A team's estimated offensive strength, according to the ODM model~\cite{govan2009} (relative to the competition)
    \item Number of shots: The total number of shots a team took during the game.
    \item Number of shots on target: The total number of shots on target a team took during the game.
    \item Number of well positioned shots: The number of shots attempted from the middle third of the pitch, in the attacking third of the pitch.
    \item Opportunities: The total number of goal scoring opportunities (as annotated in the event data).
    \item Number of attacking passes: The total number of passes attempted by a team in the attacking third of the field.
    \item Attacking pass success rate: The percentage of passes attempted in the attacking third that were successful.
    \item Number of crosses: The total number of made crosses by a team.
    \item Balls inside the penalty box: Number of actions that end in the opponents penalty box.
\end{itemize}

\noindent\textbf{Defensive strength}
\begin{itemize}
    \item Defensive strength: A team's estimated defensive strength, according to the ODM model (relative to the competition).
    \item Tackle success rate: The percentage of all attempted tackles that were successful.
\end{itemize}

\noindent\textbf{Playing style}
\begin{itemize}
    \item Tempo: The number of actions per interval.
    \item Average team position: Average x coordinate of the actions performed.
    \item Possessions: The percentage of actions in which a team had possession of the ball during the game.
    \item Pass length: The average length of all the attempted passes by a team.
    \item Attacking pass length: The average length of all attacking passes attempted by a team.
    \item Percentage of backward passes: The percentage of passes with a backward direction.
    \item Number of attacking tackles: The total number of tackles attempted in the attacking third.
    \item Attacking tackle success rate: The percentage of tackles attempted in the attacking third that were successful.
\end{itemize}

\noindent\textbf{Game situation}
\begin{itemize}
    \item Time since last goal: The number of time frames since the last scored goal.
\end{itemize}

\end{document}